\documentclass[12pt]{article}

\usepackage{amsmath}
\usepackage{natbib}

\usepackage{amsfonts}
\usepackage{amssymb}
\usepackage{graphicx}
\usepackage{color}
\usepackage{tikz}
\usepackage[colorlinks,citecolor=blue]{hyperref}
\usepackage{cite}
\usepackage[mathscr]{eucal}

\usepackage{fullpage}
\usepackage{amsthm}
\usepackage{algorithm}
\usepackage{algpseudocode}
\usepackage[margin=1in]{geometry}
\usepackage{lipsum}

\newtheorem{proposition}{Proposition}
\newtheorem{lemma}{Lemma}
\newtheorem{corollary}{Corollary}

\def\sign{\textup{sign}}
\def\diag{\textup{diag}}

\def\maximize{\textup{maximize}}
\def\R{\mathbb{R}}

\title{Supervised Classification Using Sparse Fisher's LDA}

\author{Irina Gaynanova\thanks{Corresponding author. Email: ig93@cornell.edu. Mailing address: Cornell University, Department of Statistical Science, 1173 Comstock Hall, Ithaca, NY 14853, USA.},
James G. Booth\thanks{Department of Biological Statistics and Computational Biology, Cornell University,
Ithaca,NY} and Martin T. Wells\thanks{Department of Statistical Science, Cornell University,
Ithaca,
NY}.}

\date{}

\begin{document}

\maketitle

\begin{abstract}
It is well known that in a supervised classification setting when the number of features is smaller than the number
of observations, Fisher's linear discriminant rule is asymptotically Bayes. However, there are numerous 
modern applications where classification is needed in the high-dimensional setting. Naive implementation of Fisher's rule in this case fails to provide good results because the sample covariance matrix is singular. Moreover, by constructing a classifier that relies on all features the interpretation of the results is challenging. Our goal is to provide robust classification
that relies only on a small subset of important features and accounts for the underlying correlation structure. We apply a lasso-type penalty to the discriminant vector to ensure sparsity of the solution and use a shrinkage type estimator for the covariance matrix. The resulting optimization problem is solved using an iterative coordinate ascent algorithm. Furthermore, we analyze the effect of nonconvexity on the sparsity level of the solution and highlight the difference between the penalized and the constrained versions of the problem. The simulation results show that the proposed method performs favorably in comparison to alternatives. The method is used to classify leukemia patients based on DNA methylation features. 
\end{abstract}


\section{Introduction}\label{sec:intro}

Linear discriminant analysis (LDA) is a popular method for classification when the number of observations $n$ is much bigger than the number of features $p$. If the data follows a $p$-variate normal distribution with the same covariance structure across 
the groups, LDA provides an asymptotically optimal classification rule, meaning that it converges
to the Bayes rule  \citep[Chapter~6]{Anderson:1984ty}. However, it was noted by \citet{Dudoit:2002ev} that naive implementation of LDA for
high-dimensional data provides poor classification results in comparison to alternative methods. A rigorous proof of this
phenomenon in the case $p \gg n$ is given by \citet{Bickel:2004wa}. There are two main reasons for this. First, standard LDA
uses the sample covariance matrix to estimate the covariance structure. In high dimensional settings this results in a 
singular estimator. Secondly, by using all $p$ features in classification, interpretation of the results becomes challenging.
Often, only a small subset of 
features is relevant and it is of great interest to perform classification and variable selection at the same time. 

Several methods have been proposed in the literature to account for these problems. One of the popular approaches is to use the independence rule which overcomes the singularity problem of the sample covariance but ignores the dependency structure. This approach is very appealing because of its simplicity and was encouraged by the work of \citet{Bickel:2004wa} who showed that it performs better than the standard LDA in the $p\gg n$ setting. Examples of diagonal classification methods include \citet{Tibshirani:2003bj, Fan:2008tu, Pang:2009ik, Huang:2010ju, Witten:2011kc}.

Unfortunately, independence is only an approximation and it is unrealistic in most applications. Gene interactions, for
example, are crucial for the understanding of biological processes and it is important to use this information in 
classification. Therefore we should aim for better estimators of the covariance matrix instead of using the independence
structure. Some of the main results in  \citet{Bickel:2004wa} rely critically on low rank structure of the Moore-Penrose inverse of sample covariance matrix, and so their argument can not be applied to positive definite estimators of within-group covariance matrix $\varSigma_w$. \citet[Theorem~1]{Bickel:2004wa} showed that if $p/n \to \infty$, then the misclassification rate of LDA that uses the sample covariance matrix goes to $1/2$. Their proof is based on using Moore-Penrose inverse of sample covariance matrix which has low rank. Since the diagonal estimator of $\varSigma_w$ is always positive definite, they show that independence approach is preferable in high-dimensional settings.  Indeed, in the case $\varSigma_w$ is known, the misclassification rate of independence rule is {\it always} greater than or equal to the misclassification rate of LDA that uses $\varSigma_w$. The following example illustrates that correlations are beneficial for classification if $p$ is large and the number of relevant features $(\equiv r)$ is small, in contrast to \citet{
Bickel:2004wa}.

Consider classification between two groups with the mean of the first group $\mu_1=0_{p}$ and the mean of the second group $\mu_2=\delta=(\delta_{1},...,\delta_{p})$. Also assume all the variables $x_j$ are scaled to have standard deviation one within each population. If the variables are jointly normal and equally correlated, i.e. $\mbox{cor}(x_i,x_j)=\rho$ for all $i \neq j$, then the correlation is beneficial if $\rho$ is negative or if
\begin{equation}\label{eq:rho}
 \rho>\frac{(\sum_{j=1}^p\delta_j)^2-\sum_{j=1}^p\delta_j^2}{(p-1)\sum_{j=1}^p \delta_j^2}
\end{equation}
\citep[eqn.~9]{Cochran:1964tn}.
Assume now that $\delta_j=0$ for all $j>r$ and $\delta_j=C$ for all $j\le r$, where $r\le p$. Then \eqref{eq:rho} can be rewritten as
\begin{equation}\label{eq:rho2}
 \rho>\frac{(\sum_{j=1}^rC)^2-\sum_{j=1}^rC^2}{(p-1)\sum_{j=1}^r C^2}=\frac{r-1}{p-1}.
\end{equation}
In modern high-dimensional applications it is common to assume that only small subset of features is relevant for the classification, i.e. $r<<p$. As a consequence $r/p$ is small, hence from \eqref{eq:rho2} it follows even the small values of $\rho$ are beneficial and should be accounted for in the development of discrimination rules.

We are not the first to notice that there are drawbacks to the independence approach. \citet{Huang:2010ju}
note that the discriminant scores resulting from the diagonal rule are biased. However, instead of changing the rule, they
propose adjusting for the bias directly. \citet{Fan:2010uaa} argue that it is crucial to take into account the covariance
structure. However, they avoid estimating it directly in the $p\gg n$ setting by doing preselection of features. 
\citet{Shao:2011jw} propose thresholding of the sample covariance matrix, \citet{Cai:2011dm} estimate the covariance matrix and 
mean differences directly, and \citet{Mai:2012bf} reformulate the LDA 
problem using penalized least squares. These approaches, however, are limited to the case of the two groups and the
standard LDA formulation instead of Fisher's version is considered.

The purpose of this work is to extend current methodology in a way that will enable automatic variable selection and account
for the complex dependency structure. A motivating example is the analysis of DNA methylation data from patients with Acute
Myeloid Leukemia (AML). The ERASMUS dataset was collected at Erasmus University Medical Center (Rotterdam) and consists of DNA
methylation profiles of 344 patients. The ECOG dataset is obtained from a clinical trial performed by Eastern Cooperative 
Oncology Group and consists of DNA methylation profiles of 383 patients. Samples from both cohorts were processed according
to \citet{Thompson:2008tb}. Cluster analysis performed on the ERASMUS dataset corresponded well to the available biomarker
information, providing new insights into the leukemia subtypes based on the methylation patterns (see \citet{Figueroa:2010ih}
and \citet{Kormaksson:V8tzurfm} for details). Since only limited biomarker information is available for the ECOG dataset, it
is of great interest to use the 
clustering information from the ERASMUS dataset to determine the leukemia subtypes in the ECOG dataset.

In this work we consider Fisher's formulation of LDA \citep{Mardia:1979vm}, since it is derived without explicit normality
assumption on the data. The sparsity of the solution is enforced by adding an $\ell_1$ penalty to the objective function leading to penalized FLDA. A special case of this problem has been previously considered by \citet{Witten:2011kc}, who assume diagonal covariance structure. In contrast, we propose to evaluate covariance structure through the estimator proposed by \citet{Schafer:2005in} and derive the algorithm for penalized FLDA that can be used with any other estimator of a general form. The required algorithmic modification provides additional computational challenges and we use the coordinate ascent algorithm to solve the optimization criteria. Simulation results show significant improvement in misclassification rates over the method proposed by \citet{Witten:2011kc}, especially when the true covariance structure is far from the diagonal. We also demonstrate that using a diagonal estimate of the covariance structure in penalized FLDA with two groups is equivalent to selecting features by using t-statistics with an appropriate threshold. 


In our simulation study, we noticed that in certain scenarios it is impossible to achieve a very sparse solution regardless
of the choice of the tuning parameter. We proceed by analyzing the effect of nonconvexity on the sparsity of the penalized FLDA solution and show that there is a duality gap between the $\ell_1$-penalized and the $\ell_1$-constrained formulations of the problem. With the success of the LASSO \citep{Tibshirani:1996wb}, it is very common to use the $\ell_1$ penalty as a method to achieve sparsity in the solution (for example \citet{Zou:2006ia} in PCA, \citet{Bradley:1998vq} in SVM and \citet{BIEN:vs} in covariance matrix estimation). The $\ell_1$ penalty in the LASSO is motivated by the dual problem where using the $\ell_1$ constraint geometrically means projecting the solution vector onto the subspace that forces certain components to be exactly zero. Since in the LASSO the $\ell_1$-constrained and $\ell_1$-penalized problems are equivalent, it is natural to expect that adding $\ell_1$ penalty to other objective functions provides the same effect. Unfortunately this is not always the case, due to nonconvexity, and we show that this is indeed not true in the FLDA context. We provide an intuitive explanation for this phenomenon and propose using feature clustering as a partial solution.

The rest of the paper is organized as follows. Section~\ref{sec:FLDA} reviews classical Fisher's LDA in the $n\gg p$ setting and provides the formulation of the penalized FLDA problem with the optimization details. Section~\ref{sec:nonconvexity} analyzes the effect of nonconvexity on the sparsity level of penalized FLDA solution. In Section~\ref{sec:Simulations} we compare the proposed algorithm with competing methods in simulation studies. Application to DNA methylation data is presented in Section~\ref{sec:data}. We conclude in Section~\ref{sec:discus} with a discussion.

\section{A Penalized FLDA}\label{sec:FLDA}
\subsection{Fisher's Linear Discriminant Analysis (FLDA)}\label{sec:method}
Let $X\in\R^{n\times p}$ be the data matrix where observations $X_{i}\in\R^{p}$, $i=1,...,n$, are independent and come from one of the $g$ groups with the same within-group covariance matrix $\varSigma_{w}$. The between-group covariance matrix $\varSigma_b$ is defined as $\varSigma_{b}=\sum_{i=1}^{g}\pi_{i}(\mu_{i}-\mu)(\mu_{i}-\mu)^{T}$, where $\pi_{i}$ is the prior probability that observation comes from group $i$, $\mu_{i}$ is the mean for the $i$th group and $\mu=\sum_{i=1}^{g}\pi_{i}\mu_{i}$ is the overall population mean.

 Consider the within-group sample covariance matrix $W=\frac1{n}\sum_{i=1}^{g}n_{i}S_{i}$ and the between-group sample covariance matrix $B=\frac1{n}\sum_{i=1}^{g}n_{i}(\bar X_{i}-\bar X)(\bar X_{i}-\bar X)^{T}$, where $n_i$ is the number of observations in the $i$th group, $S_{i}$ is the sample covariance matrix for the $i$th group,  $\bar X_{i}$ is the sample mean for the $i$th group and $\bar X$ is the overall sample mean. 

Fisher's linear discriminant analysis (FLDA) \citep{Mardia:1979vm} seeks linear combinations $v_{r}\in \R^{p}$, $r=1,...,g-1$, of $p$ features such that
\begin{align*}
v_{r}=\arg\max_{v} v^{T}Bv \quad \mbox{subject to} \quad  v^TWv=1\mbox{, }v^{T}Wv_{j}=0\mbox{ for }j<r.
\end{align*}
The vectors $v_{r}$ that solve this problem are the eigenvectors corresponding to the non-zero eigenvalues of the rank $(g-1)$ matrix $W^{-1}B$. Therefore the aim of FLDA is to find linear combinations of features that maximize the between-group variability with respect to the within-group variability.

Let $V=(v_{1}....v_{g-1})\in \R^{p\times(g-1)}$. Then FLDA  classifies a new observation $X\in \R^{p}$ with the value $x$ according to the rule $\hat h(x)$, where
\begin{equation*}
\hat h(x)=\arg\min_{i=1,...,g}(x-\bar X_{i})^{T}V^{T}V(x-\bar X_{i}).
\end{equation*}

Further we focus on the problem of estimating the leading discriminant vector $v=v_{1}$:
\begin{equation}\label{eq:v}
\begin{split}
v=\arg\max_{v} v^{T}Bv \quad \mbox{subject to}\quad v^TWv=1.
 \end{split}
\end{equation}
 We discuss how to obtain multiple discriminant vectors in Appendix~A.3.

\subsection{Penalization of FLDA}\label{sec:penFLDA}
 When $p>>n$, $W$ is singular, hence the matrix $W^{-1}B$ is not well-defined. Moreover, it is usually expected that only a relatively small subset of $p$ features is relevant for the discriminant rule. Optimization problem~\eqref{eq:v}, however, results in $v$ with only non-zero components, effectively using all $p$ features. The extension of FLDA to high-dimensional settings requires finding solutions to both of these problems.
 
In this article we consider the penalized Fisher's linear discriminant problem:
\begin{equation}\label{eq:LDAwithinTilde}
v_{\lambda}=\arg\max_{v\in \R^{p}}\left \{v^{T} Bv-\lambda\sum_{j=1}^{p}|s_{j}v_{j}|\right \} \mbox { subject to } v^{T}\tilde Wv\le1.
\end{equation}
Here $s_{j}$ is the within-group standard deviation for feature $j$ and $\tilde W$ is a positive definite estimator of $\varSigma_{w}$. As in the LASSO \citep{Tibshirani:1996wb}, the $\ell_1$ penalty in the objective function of \eqref{eq:LDAwithinTilde} leads to sparse solution $v_{\lambda}$. 

We propose to use Algorithm~\ref{a:nondiag} to solve \eqref{eq:LDAwithinTilde}, which is a variation of the Alternate Convex Search (ASC) \citep[Section~4.2.1]{Gorski:2007uv}. Here $S$ is a soft-thresholding operator, i.e. $S(x,a)=\sign(x)(|x|-a)_{+}$. 
The full derivation of the algorithm and optimization details are presented in Appendix~A. Instead of doing a linear update of all $p$ features, a random update is performed through the use of $\tilde L\gets\mbox{sample}(1,...,p)$. This guarantees a faster convergence rate of the coordinate ascent method \citep{ShalevShwartz:2009kl}.

\begin{algorithm}[!h]
Given: $\lambda>0$, $B$, $\tilde W$, $k=1$
\begin{algorithmic}
\State $v^{(0)} \gets\mbox{ dominant eigenvector of }\tilde W^{-1}B$
\State $v^{(0)} \gets v^{(0)}/\sqrt{(v^{(0)})^{T}\tilde W v^{(0)}}$
\State $\bar q \gets v^{(0)}$
\Repeat
\Repeat
	\State $D\gets 0$
	\State $\tilde L\gets\mbox{sample}(1,...,p)$
	\For{$l\in\tilde L$}
 	 \State $q_{l} \gets S\left((Bv^{(k-1)})_{l}-\sum_{i\neq l}w_{li}\bar q_{l},\lambda s_{l}/2\right)/w_{ll}$
	 \State $D=D+|q_{l}-\bar q_{l}|$
	 \State $\bar q_{l}=q_{l}$
	\EndFor
\Until{$D<\epsilon$}	
\If{$\bar q \neq 0$}\State $v^{(k)}\gets \bar q/\sqrt{\bar q^{t}\tilde W\bar q}$
\Else \State $v^{(k)}\gets \bar q$
\EndIf
\State $k\gets k+1$
\Until{$k=k_{\max}$ or $v^{(k)}$ satisfies stopping criterion.}
\end{algorithmic}
\caption{Optimization algorithm for penalized FLDA.}
\label{a:nondiag}
\end{algorithm}

\citet{Witten:2011kc} have considered  a special case of problem~\eqref{eq:LDAwithinTilde} with $\tilde W=\diag(s_{j}^{2})$, effectively assuming the independence of features. This assumption significantly simplifies the update of vector $q$ in Algorithm~\ref{a:nondiag}, leading to a single update for each $l$th feature
\begin{equation}\label{eq:kupdateWitten}
q_{l}=\frac{S\left((Bv^{(k-1)})_{l},\frac{\lambda s_l}2\right)}{w_{ll}}.
\end{equation}
As a result, the choice of diagonal estimator $\tilde W=\diag(s_{j}^{2})$ has a strong computational advantage, although this model is clearly misspecified.

In Section~\ref{sec:intro} we reviewed the importance of correlations in classification. Therefore, instead of assuming independence, we propose to use the following estimator:
\begin{equation*}
\tilde W=\sum_{i=1}^g n_i\tilde S_i,
\end{equation*}
where $\tilde S_i=\tau_i \diag(S_i)+(1-\tau_i) S_i$  \citep{Schafer:2005in} and $S_i$ is the sample covariance matrix for the $i$th group. There are
several advantages to using this estimator. First, the resulting matrix $\tilde W$ is always positive definite and it
preserves the correlation structure of the data. Secondly, an optimal $\tau_i$ can be chosen according to \citet{Ledoit:2004hu}
that guarantees minimal MSE under the existence of the first two moments. Therefore there are no strong distributional
assumptions on the data.  Finally, given the simple form of the estimator, $\tilde W$ can be computed easily and quickly
for the large values of $p$. This approach is similar to the one used in regularized discriminant analysis
\citep{Friedman:1989tm, Guo:2007te}. Aside from the feature selection procedure, our approach is different in that each
within-group covariance matrix is shrunk towards a diagonal estimate instead of the identity matrix, and the shrinkage parameter $\tau_i$ is selected automatically.

Algorithm~\ref{a:nondiag} can be used for any positive definite estimator $\tilde W$ and therefore allows to use alternative estimators of the covariance structure $\varSigma_w$. Often it is desirable for the estimate to be both positive definite and sparse. Here by sparse we mean that certain entries of the covariance matrix are estimated exactly as zero. There is an extensive literature on covariance matrix estimators that achieve one of these goals. However, to our knowledge only limited methodology is available for achieving both. Among recent advances are methods proposed by \citet{BIEN:vs} and  \citet{ROTHMAN:wi}. Unfortunately, the estimators considered by these authors are very computationally intensive and do not scale well to high-dimensional data sets.

We proceed by analyzing the variable selection properties of $v_{\lambda}$ in~\eqref{eq:LDAwithinTilde} and derive a surprising result that there is a lower bound on the number of non-zero features in $v_{\lambda}$ (Corollary~\ref{cl:mlambda}). The existence of a lower bound implies that the method can not always provide very sparse solutions, which challenges the interpretation of the results when $p$ is large. To overcome this problem, we introduce feature clustering in Section~\ref{sec:clustering}.

\section{Effect of nonconvexity on the sparsity level of penalized FLDA solution}\label{sec:nonconvexity}

\subsection{Empirical evidence}
\begin{figure}
\centering
\includegraphics[scale=0.5]{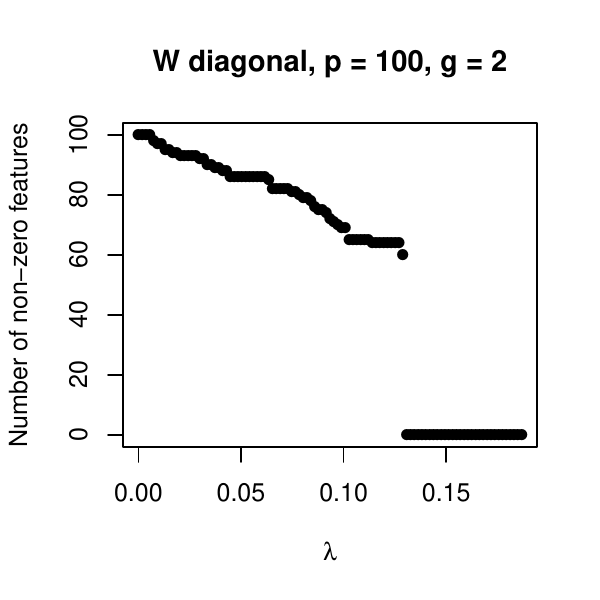}
\includegraphics[scale=0.5]{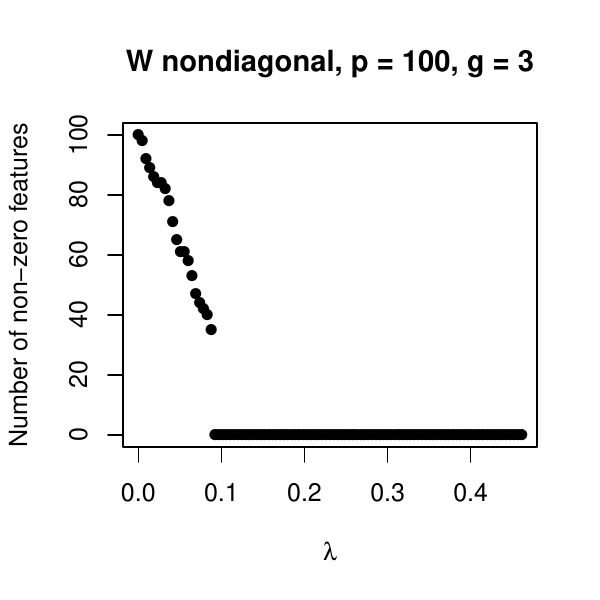}
\caption{Number of non-zero features versus tuning parameter $\lambda$.}
\label{fig:drop1}
\end{figure}

The larger the value of $\lambda$ in \eqref{eq:LDAwithinTilde}, the larger the penalty on $v$ and therefore the smaller the number of non-zero components in $v_{\lambda}$.  In our simulations, we noticed that there seemed to be no value of $\lambda$ that resulted in a small number of non-zero components. As illustrated in Figure~\ref{fig:drop1}, this behavior is observed for different values of $g$ and different estimators $\tilde W$.

Figure~\ref{fig:drop1} indicates that the sparsity level of $v_{\lambda}$ is not smooth in $\lambda$. Specifically, there exists a $\lambda_{0}$ such that for all $\lambda<\lambda_{0}$ the solution $v_{\lambda}$ has at least $M$ non-zero components and for all $\lambda\geq\lambda_{0}$ $v_{\lambda}=0$. To understand the origins of this behavior, we contrast the $\ell_{1}$-penalization with $\ell_{1}$ constraint and demonstrate the effect of nonconvexity using geometrical arguments (Section~\ref{s:geometry}). We also derive a lower bound on the number of non-zero features in $v_{\lambda}$ for the case $g=2$ and $\tilde W=\diag(s_{j}^{2})$ (Section~\ref{s:lowerbound}). While Figure~\ref{fig:drop1} indicates the existence of such a bound in the case $g>2$ and nondiagonal $\tilde W$, the additional complexity of the optimization problem~\eqref{eq:LDAwithinTilde} makes the theoretical analysis very challenging. 

\subsection{$\ell_{1}$-penalization versus $\ell_{1}$ constraint}\label{s:geometry}

In Section \ref{sec:FLDA} we considered $\ell_1$-penalized Fisher's LDA, where the $\ell_1$ norm was incorporated into the objective function
\begin{equation}\label{eq:LDAl}
v_{\lambda}=\arg\max_{v \in \mathbb{R}^{p}} \left\{ v^{T}Bv-\lambda\|v\|_1 \right\} \mbox{ subject to } v^{T} \tilde Wv\le 1.
\end{equation}
We omit the additional penalty weights for ease of demonstration.

A closely related problem can be formulated by incorporating an $\ell_1$ norm directly in the constraint set:
\begin{equation}\label{eq:LDAt}
v_{t}=\arg\max_{v\in \mathbb{R}^{p}} v^{T}Bv\mbox{ subject to } v^{T} \tilde Wv\le 1, \|v\|_1 \le t,
\end{equation}
where $t \ge 0$ is a tuning parameter. Solving problem \eqref{eq:LDAt} is computationally more challenging than solving problem \eqref{eq:LDAl}. Optimization problem \eqref{eq:LDAt} can be solved using Algorithm~\ref{a:nondiag} with the following modification: instead of using a fixed $\lambda$ for each iteration $k$,  one needs to find $\lambda^{(k)}$ such that $\|v^{(k)}_{\lambda^{(k)}}\|_{1}=t$. Usually, such a $\lambda^{(k)}$ is found by performing a binary search on the grid $[0,\lambda_{\max}]$ for each iteration $k$.
\begin{proposition}\label{cl:dualone}
For every $\lambda \ge 0$ there exists $t \ge 0$ such that $v_{\lambda}=v_{t}$.
\end{proposition}
The reverse, however, is not generally true. In the LASSO setting \citep{Tibshirani:1996wb}, the optimization problem is convex and therefore the solutions to $\ell_1$-penalized and $\ell_1$-constrained optimization problems coincide \citep[Proposition~5.2.1]{Bertsekas:1999ua}. Unlike the LASSO, \eqref{eq:LDAt} is not a convex problem and therefore this guarantee no longer applies. Because of the form of the constraint set in~\eqref{eq:LDAt}, the sparsity level of $v_{t}$ changes smoothly with $t$. However, since not every solution to \eqref{eq:LDAt} is the solution to \eqref{eq:LDAl}, such behavior is not guaranteed for $v_{\lambda}$ and $\lambda$. 

Consider the case $p=2$, $g=2$ and $\tilde W=I$. It follows that $B$ is a $ 2 \times 2$ matrix with its rank equal to one. Therefore it is completely determined by its positive eigenvalue $\gamma$ and the corresponding eigenvector $l$: $B=\gamma ll^{T}$. We set $\gamma=1$ and consider two scenarios: in the first scenario, we initialize the eigenvector $l$ as $l=(0.2,0.8)^{T}$ and in the second scenario as $l=(0.5, 0.6)^{T}$. In both cases $l$ is then normalized to have $\ell_2$ norm exactly equal to one, $l^{T}l=1$. 

\begin{figure}
\centering
\includegraphics[scale=0.27]{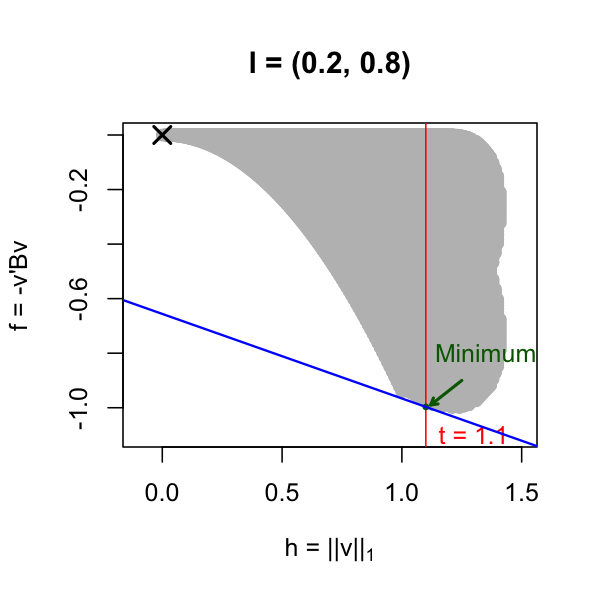}
\includegraphics[scale=0.27]{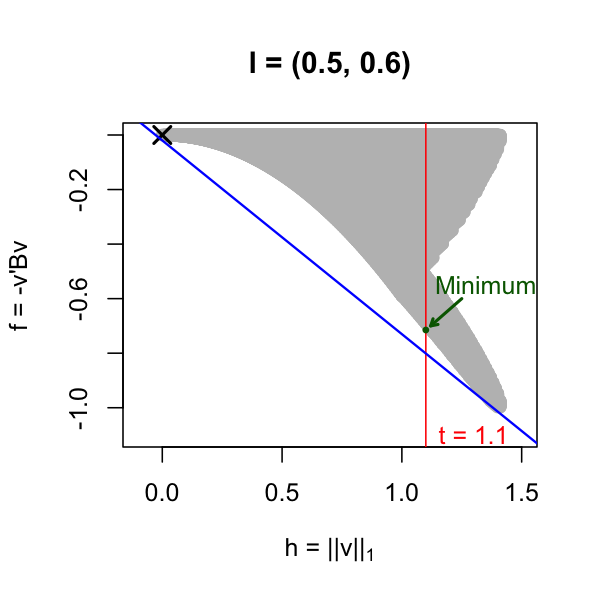}
\caption{\label{fig:setS2}Visualization of the set $S$ and the minimum common point of $S$ and $h\le t$, $l$ is the eigenvector of matrix $B$ before normalization. The $(0,0)$ point is indicated with \textbf{x}. Blue line is the supporting hyperplane for the set $S$.}
\end{figure}

Consider the corresponding $\ell_{1}$-constrained optimization problem: 
\begin{equation}\label{eq:minLDAt}
v_{t}=\arg\min_{v\in \mathbb{R}^{p}} -(v^{T}l)^{2}\mbox{ subject to } v^{T}v\le 1, \|v\|_1 \le t.
\end{equation}
Following \citet[Chapter~5]{Bertsekas:1999ua} and \citet[Chapter~5.3]{Boyd:2004uz}, we use a geometry-based approach to visualize the relationship between the solutions to $\ell_1$-constrained and $\ell_1$-penalized optimization problems. Consider set $S$ of constrained pairs 
\begin{equation}\label{eq:S}
S=\left\{(h,f)\left|h=\|v\|_{1}, f=-(v^{T}l)^{2}\mbox{ for all } v\in \mathbb{R}^{p}, v^{T}v\le 1 \right.\right\}.
\end{equation}
Note that problem \eqref{eq:minLDAt} can be formulated as a minimal common point problem: finding a point with a minimal $f$th coordinate among the points common to both set $S$ and halfspace $h\le t$. In other words, finding a point $(h',f')$ such that
\begin{equation}\label{eq:hf1}
\left\{(h',f')\in S\left|f'=\min_{(h,f)\in S, h\le t}{f}\right.\right\}.
\end{equation}
By definition of $v_{t}$ it follows that $f'=-(v_{t}^{T}l)^{2}$ and $h'=\|v_{t}\|_{1}$. We construct the corresponding sets $S$ for both scenarios in Figure \ref{fig:setS2} and identify the minimal common point for each of them using value of $t=1.1$.

Consider the corresponding $\ell_{1}$-penalized optimization problem:
\begin{equation}\label{eq:minLDAl}
v_{\lambda}=\arg\min_{v\in \mathbb{R}^{p}} -(v^{T}l)^{2}+\lambda\|v\|_{1}\mbox{ subject to } v^{T}v\le 1.
\end{equation}
Using the set $S$ in \eqref{eq:S} we can reformulate this minimization problem as finding the point $(h'',f'')\in S$ such that
\begin{equation*}
(h'',f'')=\arg \min_{(h,f)\in S} \{f+\lambda h\}.
\end{equation*}
It follows that $h''=\|v_{\lambda}\|_{1}$ and  $f''=-(v_{\lambda}^{T}l)^{2}$. Note that the solutions to \eqref{eq:minLDAt} and \eqref{eq:minLDAl} are the same if $h''=h'$ and $f''=f'$ with $(h',f')$ from~\eqref{eq:hf1}. These points are the same when $f=-\lambda h$ is the supporting hyperplane for the set $S$ at the point $(h',f')$. In Figure \ref{fig:setS2} we evaluate whether such a hyperplane can be constructed in both scenarios. In the first scenario such a hyperplane can be constructed for each $t\ge 1$, in particular for $t=1.1$. In the second scenario, such a hyperplane can not be constructed for $t=1.1$ as it has to lie below the point $(0,0)$ and the minimal point of $S$ corresponding to $h=1.4$. Moreover, this is true not only for $t=1.1$ but for all values of $t$ between $1$ and $1.4$. Hence for these $t$ there exists no $\lambda$ such that $v_{\lambda}=v_{t}$. 

The shape of the set $S$ in the second scenario suggests an interesting implication on the sparsity of the solution $v_{\lambda}$. For all $t<1.4$ the only point $(h',f')$ from~\eqref{eq:hf1} at which we can construct the supporting hyperplane  to the set $S$ is the point $(h',f')=(0,0)$. This implies $\|v_{\lambda}\|_{1}=0$, hence $v_{\lambda}=0$ is the corresponding solution to the dual problem \eqref{eq:minLDAl} for all $t<1.4$. However, $v_{t}=0$ only for $t=0$.  Therefore there exists no $\lambda\ge 0$ such that $\|v_{\lambda}\|_{1}=t$ for $t\in(0,1.4)$. Hence, there is a constraint on the sparsity level of the solution $v_{\lambda}$ and a sudden drop from $\|v_{\lambda}\|_{1}=1.4$ to $\|v_{\lambda}\|_{1}=0$ can be observed as $\lambda$ increases. This is consistent with Figure~\ref{fig:drop1}.

In the language of optimization theory, the Lagrangian dual problem defines the supporting hyperplane to $S$ in~\eqref{eq:S}, hence the optimal (primal) solution is greater than the dual solution (weak duality). If the supporting hyperplane intersects $S$ at a single point, as in scenario one above, the optimization problem is said to have the zero duality gap (strong duality) property.  If the objective function is convex, as in the LASSO, strong duality is guaranteed by SlaterÕs constraint \citep[Chapter~5]{Boyd:2004uz}.

\subsection{Lower bound on the number of non-zero features in $v_{\lambda}$}\label{s:lowerbound}
In this subsection we derive a lower bound on the number of non-zero features in $v_{\lambda}$ under two simplifying assumptions: $g=2$ and $\tilde W=\diag(s_{j}^{2})$. While our main goal is to quantify the drop in the number of non-zero features observed in Figure~\ref{fig:drop1}, we also gain new insights into the inferential methodology proposed by \citet{Witten:2011kc}.  In particular, the choice of $\tilde W=\diag(s_{j}^{2})$ leads to selecting features with the largest absolute values of the t-test statistics $t_{i}=(\bar X_{1i}-\bar X_{2i})/s_{i}$ (Proposition~\ref{cl:ttest}).

For our derivations it is sufficient to consider the case $\tilde W=I$. Indeed, consider a new vector $z\in \mathbb{R}^{p}$ such that $z_j=v_j s_j$ for $j=1,..,p$ or equivalently $z=\tilde W^{1/2}v$.  Problem \eqref{eq:LDAwithinTilde} can be rewritten as $v_{\lambda}=\tilde W^{-1/2}z_{\lambda}$ with
\begin{equation}\label{eq:zLDAs1}
z_{\lambda}=\arg \max_{z \in \mathbb{R}^{p}} z^{T}\tilde W^{-1/2}B\tilde W^{-1/2}z-\lambda\sum_{j=1}^p|z_j| \mbox{ subject to } z^{T}z\le 1.
\end{equation}
Since $\tilde W$ is diagonal, $v_{\lambda}$ and $z_{\lambda}$ have the same sparsity pattern. Let $t$ be a vector of t-test statistics, $t=(t_{1},...,t_{p})$. From definition of $B$ it follows that $B'=\tilde W^{-1/2}B\tilde W^{-1/2}\propto tt^{T}$ and $z^{T}\tilde W^{-1/2}B\tilde W^{-1/2}z \propto (z^{T}t)^{2}$. This observation leads to the following result.
\begin{proposition}\label{cl:ttest} Assume $g=2$ and $v_{\lambda}$ is found as solution to \eqref{eq:LDAwithinTilde} with $\tilde W=\diag(s_{j}^{2})$. Define $t_{i}=(\bar X_{1i}-\bar X_{2i})/s_{i}$, $i=1,\dots,p$. Then for each $\lambda>0$ there exists $C_{\lambda}>0$ such that
$$
\{i:v_{\lambda i}\neq 0\}=\{i:|t_{i}|>C_{\lambda}\}.
$$
\end{proposition}

\textit{Remark:}
\citet[Section~7.2]{Witten:2011kc} indirectly explore the connection with the t-test statistics through connection with nearest shrunken centroids \citep{Tibshirani:2003bj}. Their result, however, requires modification of the objective function in optimization problem~\eqref{eq:LDAwithinTilde} ($\sqrt{v^{T}Bv}$ instead of $v^{T}Bv$ is used) and is limited to the case $n_{1}=n_{2}$. Moreover, the statement and proof of Proposition~4 in \citet{Witten:2011kc} concern the classification rule, whereas we focus on the variable selection.

Let $l$ be the dominant eigenvector of $B'$, normalized as $l^{T}l=1$. Since $B'\propto tt^{T}$, the largest values of $|l_{i}|$ correspond to the largest values of $|t_{i}|=|\bar X_{1i}-\bar X_{2i}|/s_{i}$. Without loss of generality we assume that the features are ordered so that $|t_{1}|\ge |t_{2}|\ge....\ge |t_{p}|>0$. Indeed, if $|t_j|=0$ for some $j$, then from \eqref{eq:zLDAs1} it follows that $z_{\lambda j}=0=v_{\lambda j}$ for any $\lambda>0$. For each $j=0,...,p$ we define the vector $l^{j}\in \mathbb{R}^{p}$ with $i$th component 
 $$
 l^{j}_i=\begin{cases}
       l_{i} &\text{if $i \leq j$};\\
       0, &\text{if $i>j$}.
      \end{cases}
 $$
In other words, $l^{j}$ is equal to eigenvector $l$ truncated at the $(j+1)$th component, with $l^{0}=0$ and $l^{p}=l$. We also define the sets $A_{j}$ such that 
\begin{equation*}
\begin{split}
A_j&=\{z \in \mathbb{R}^{p}: \mbox{first $j$ components of vector $z$ are non-zero, all other components are zero}\}\\
&=\{z \in \mathbb{R}^{p}: \|z\|_{0}=j; \mbox{ }z_{j+1}=...=z_{p}=0\}.
\end{split}
\end{equation*}

Given this definition and the ordering of the test statistics $t_{i}$, Proposition~\ref{cl:ttest} implies that for each $\lambda$ the solution $z_{\lambda}$ belongs to one of the sets $A_{j}$,  $z_{\lambda}\in \bigcup_{j=0}^{p} A_{j}$. We further show that for each $\lambda>0$ there exists $m_{\lambda}\ge 1$ such that $z_{\lambda}\in A_0\cup \bigcup_{i=m_{\lambda}}^{p}A_i$ (Proposition~\ref{cl:additional}). Thus, depending on the $\lambda$, the solution to \eqref{eq:zLDAs1} has at least $m_{\lambda}$ non-zero features or is exactly 0. We also show that there exists $m\ge 1$ such that $z_{\lambda}\in A_0\cup \bigcup_{i=m}^{p}A_i$ for any $\lambda>0$ (Proposition~\ref{cl:main}).

\begin{proposition}\label{cl:additional}
Let $m_{\lambda}=j_{\min}$ such that $\|l^j\|_2>\lambda/(\gamma |l_1|)$, $j=1,...,p$. Then $$z_{\lambda}\in A_0\cup \bigcup_{i=m_{\lambda}}^{p}A_i.$$ If there is no such $j$, then $z_{\lambda}=0$.
\end{proposition}
\noindent
\textit{Remark:}
For $\lambda <\gamma |l_1|$, the value of $m_{\lambda}$ increases with $\lambda$.  If $\lambda \ge \gamma |l_1|$ then $z_{\lambda}=0$.
\begin{proposition}\label{cl:main} Let $m=j_{\max}$, $j=1,...,p$, such that there exists $r\ge j$ with 
$$\|l^{j-1}\|_2|l_1|\le \frac{\|l^r\|^{3/2}}{\|l^r\|_1}.$$
 Then $v_{\lambda} \in A_0\cup \bigcup_{i=m}^p A_i$ regardless of the choice of the tuning parameter $\lambda$.
\end{proposition}

\begin{corollary} \label{cl:mlambda}
Let $g=2$ and $\tilde W=\diag(s_{j}^{2})$. The solution $v_{\lambda}$ to problem \eqref{eq:LDAwithinTilde} is either zero or has $M$ non-zero components, where
$$
M\ge M_{\lambda}=\max(m_{\lambda},m)
$$
and $m_{\lambda}$, $m$ are defined in Propositions~\ref{cl:additional} and~\ref{cl:main}, respectively. 
\end{corollary}

\begin{figure}
\centering

\includegraphics[scale=0.5]{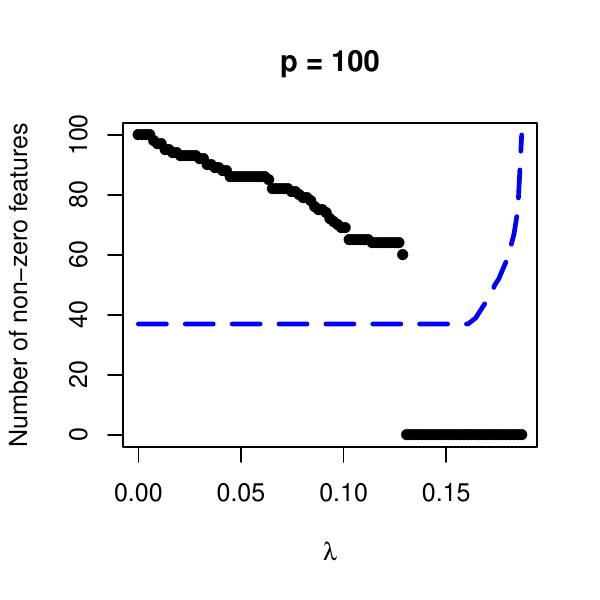}
 \includegraphics[scale=0.5]{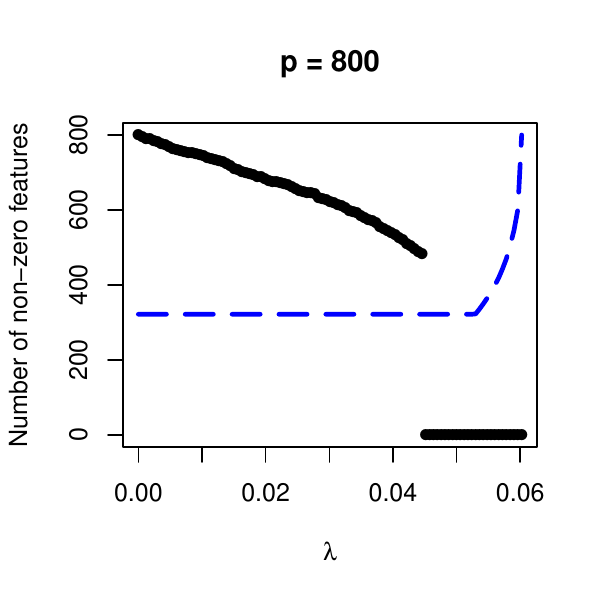}

\caption{\label{fig:bounds}Number of non-zero features obtained empirically (dots) and the theoretical threshold $M_{\lambda}$ (dashed line) versus the tuning parameter $\lambda$.}
\end{figure}

 We further investigate via simulations how the value of $M_{\lambda}$ varies with $\lambda$ and $p$. We also assess how close $M_{\lambda}$ is to the number of non-zero features obtained empirically. For this purpose we generate $l$ as a random vector with each component $l_i$ coming from the uniform distribution on $[0,1]$. We then standardize $l$ to have $\ell_2$ norm equal to one and order the features so that $|l_1|\ge|l_2|\ge...\ge|l_p|>0$. Finally, we solve \eqref{eq:zLDAs1} for the range of tuning parameter $\lambda$ and compare the resulting number of non-zero features with $M_{\lambda}$. The results are shown in Figure~\ref{fig:bounds}. 

The simulations indicate that the value of $M_{\lambda}=\max(m_{\lambda},m)$ increases both with the value of $\lambda$ and with the value of $p$. Note that the increase in $\lambda$ is due to $m_{\lambda}$ increasing with $\lambda$, hence $m_{\lambda}>m$ for the large values of $\lambda$. On the other hand, $m>m_{\lambda}$ for the small values of $\lambda$, which explains the dashed line being initially parallel to the x-axis in Figure~\ref{fig:bounds}. Since for both $p=100$ and $p=800$ the features $l_i$ were generated uniformly at random, the observed relationship between $M_{\lambda}$ and $p$ can not be explained by the structural differences in $l$. However, $l$ is restricted to have $\ell_2$ norm equal to one, which means that the increase in $p$ leads on average to a decrease in the differences between the individual features of $l$. This observation suggests that the larger the differences between the features of $l$, the higher level of sparsity that can be achieved by solving $\ell_1$-penalized problem \eqref{eq:LDAwithinTilde}. In Section \ref{sec:clustering} we show that the theoretical bound $m$ from Proposition \ref{cl:main} is consistent with this intuition.

\subsection{Clustering}\label{sec:clustering}

In this subsection we provide an interpretation of the conditions of Proposition \ref{cl:main} , which leads us to propose feature clustering as a way to get very sparse solutions. From Proposition~\ref{cl:main}, the larger the value of $m$, the less sparse are obtained solutions. Specifically, a large value of $m$ corresponds to a large value of $j$ for which there exists $r\geq j$ with 
\begin{equation}\label{eq:lcond}
\|l^{j-1}\|_2|l_1| \le \frac{\|l^r\|^{3/2}_{2}}{\|l^r\|_1}.
\end{equation}
Inequality~\eqref{eq:lcond} can be rewritten as 
$\|l^{j-1}\|_2 \|l^r\|_1|l_1|\le \|l^r\|_2\|l^r\|^2_2.$
Given the ordering of the features of the vector $l$, $|l_{1}|\ge...\ge|l_{p}|$, and the fact that $r\geq j$, it is always the case that
$\|l^{j-1}\|_2< \|l^r\|_2$ and $\|l^r\|_1|l_1|\ge \|l^r\|^2_2.$ Hence, for~\eqref{eq:lcond} to hold, the difference between $\|l^r\|_{1}|l_1|$ and $\|l^r\|^2_2$ should be relatively small in comparison to the difference between $\|l^{j-1}\|_2$ and $\|l^r\|_2$. In particular, in the extreme case when $|l_{1}|=...=|l_{r}|$, $\|l^{r}\|_{1}|l_1|=\|l^{r}\|_{2}^{2}$, and~\eqref{eq:lcond} is equivalent to $\|l^{j-1}\|_2\leq \|l^r\|_2$, which is always true for $r\geq j$ leading to $m=p$. It follows that the closer the features of the vector $l$ are to each other, the more likely~\eqref{eq:lcond} holds leading to a large value of $m$.

The two scenarios described in Section~\ref{s:geometry} are consistent with this intuition. The difference between the two scenarios is only in the matrix $B$, or more precisely, in how close the features of the eigenvector $l=(l_{1},l_{2})^{T}$ are to each other. The absolute difference $|l_{1}-l_{2}|$ is smaller in the second scenario ($0.1$ versus $0.6$ before normalization). 
Proposition~\ref{cl:main} gives $m=1$ for the first scenario and $m=2$ for the second. 

Recall that when $g=2$, the eigenvector $l\propto t$, where $t$ is a vector of t-test statistics. This observation leads to a more intuitive interpretation of Proposition~\ref{cl:main}. If some features have similar values of the mean differences with respect to their standard deviation, then the penalized criterion \eqref{eq:zLDAs1} either selects the whole group of features or eliminates the whole group. 

In case a very sparse solution is desired, we propose to use t-test statistics to cluster the features and create new meta-features as the averages over features belonging to each cluster.  The algorithm for $\ell_{1}$-penalized criterion can then be applied to the meta-features. There are several advantages to this procedure. First, the dimensionality of the original problem is reduced from $p$ to the number of clusters $k$. Secondly, the procedure is justifiable from a biological perspective, since it reflects a reluctance to choose only one feature among the group of features that have very similar behavior. Finally, the differences between the meta-features are larger than the differences between the original features, which helps to overcome the problem of finding a sparse solution discussed above.

\section{Simulation results}\label{sec:Simulations}
The following classification methods are considered for comparison: penalized FLDA proposal of \citet{Witten:2011kc} with $\tilde W=\diag(s_{j}^{2})$; penalized FLDA with nondiagonal $\tilde W$ found according to Section~\ref{sec:penFLDA}; sparseLDA proposal of \citet{Clemmensen:2011kr} and DSDA proposal of \citet{Mai:2012bf}. In what follows, we refer to these methods as FLDAdiag, FLDA, sparseLDA and DSDA. For the simplicity of illustration, the comparison is performed for the case of the two groups assuming the data comes from the multivariate normal distribution. In all cases, the tuning parameter is selected by 5-fold cross-validation.
The following covariance structures are considered:
\begin{enumerate}
\item Diagonal unit variance, $\varSigma_w=I$
\item Blockdiagonal with a network structure. We randomly choose positions for 40 blocks each of size 4. Within each block the offdiagonal elements are set to 0.75. Five pairs of blocks were chosen randomly to have a correlation of 0.7 between their elements. This initialization is an attempt to mimic a network structure in which there is strong correlation between elements that are close to one another in addition to strong correlation between certain groups of elements that are not necessarily close. Our motivation for considering this structure is based on \citet{Xiao:2011iw}, who consider spatial correlations between genes linearly separate on the chromosome but spatially close with respect to its three-dimensional structure.
\item Based on DNA methylation data.
The covariance matrix $\varSigma_w$ is estimated according to \citet{Schafer:2005in} based on the features selected from the ERASMUS DNA methylation dataset. 
\end{enumerate}

The clustering idea described in Section \ref{sec:clustering} is not implemented in the simulation studies to make the comparisons between different methods easier. The training set has 100 observations per group and the test set has 500 observations per group. The dimension $p$ is set to 800. At each iteration, the training set and the test set are generated from a multivariate normal distribution with means $\mu_1$ and $\mu_2$ and covariance structure $\varSigma_w$. The mean of the first group is set to $\mu_1=0$, while the mean of the second group is non-zero for the first $r=80$ features (with values ranging from 0.2 to 0.6) and 0 everywhere else.  This configuration reflects the $p\gg n$ framework with only $10\%$ relevant features.

\textit{Remark:}
The choice of $\mu_2$ is  influenced by the Example 11.8.1 from \citet{Mardia:1979vm}, where it is shown that the correlation is especially influential in classification when there is a difference in the components of the mean vector. 


The simulation results are summarized in Table \ref{tab:simul}. Errors correspond to average percent of misclassified observations; Features show how many variables on average were selected by each algorithm; and Correct Features indicate how many features were selected out of original $r=80$. 
\begin{table}

\centering
\caption{Comparison of FLDA, FLDAdiag and sparseLDA over 25 iterations, mean values are reported with standard deviation given in brackets}
\begin{tabular}{llcccc}
 \\
&Covariance matrix & FLDA & FLDAdiag & sparseLDA & DSDA \\[5pt] 
Error&Diagonal & 6.92(1.13) & 7.26(1.26) & 12.4(2.23) & 11.89(1.69) \\ 
  (in percentage)&Blockdiagonal & 19.41(2.07) & 20.34(2.15) & 21.87(2.95) & 20.97(2.09) \\ 
 & Data Based & 3.57(0.75) & 30.55(2.32) & 10.17(1.26) & 7.03(1.2) \\[3pt] 
  Features&Diagonal & 230(117) & 244(87) & 59(19) & 93(24) \\ 
  &Blockdiagonal & 271(122) & 359(211) & 38(21) & 51(19) \\ 
  &Data Based & 415(110) & 172(141) & 95(7) & 181(13) \\[3pt] 
 Correct &Diagonal & 70(4) & 71(5) & 40(6) & 47(4) \\ 
 Features &Blockdiagonal & 71(5) & 73(6) & 22(6) & 26(5) \\ 
  &Data Based & 71(4) & 69(7) & 32(3) & 47(4) \\ 
\end{tabular}
\label{tab:simul}
\end{table}
Since FLDAdiag explicitly uses the diagonal estimate of the covariance structure, we expected it to be the best among the methods for the case $\varSigma_w=I$. However, the error rates of the FLDA and FLDAdiag are the same. DSDA and sparseLDA perform much worse for this covariance structure. For the blockdiagonal structure the classification performance of all the methods is comparable. The most significant difference between the methods is observed for the covariance structure estimated from the real data. In this case FLDAdiag has much higher misclassification rate than the competitors, with FLDA demonstrating the best performance in terms of error rates.

\begin{figure}
\centering
\includegraphics[scale=0.6]{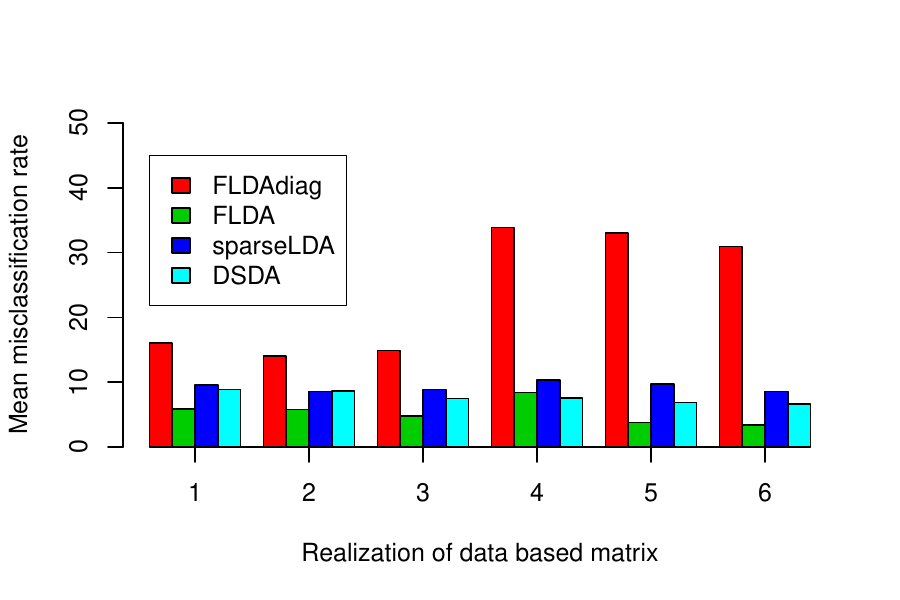}
\caption{\label{fig:simcomp}Comparison of methods on data based covariance matrices.}
\end{figure}

To make sure that the poor performance of FLDAdiag on the Data Based structure is not affected by the particular choice of 800 features that were used to generate the covariance matrix, we independently generate 6 additional covariance structures: 3 from ERASMUS DNA methylation dataset and 3 from ERASMUS gene expression dataset. Each structure is generated using a different set of 800 features and the prediction error is calculated based on the 10 iterations. The results are presented in Figure~\ref{fig:simcomp}. Note that FLDA performs consistently better than FLDAdiag with approximately a 3-fold reduction in classification errors.

\begin{figure}
\centering
\makebox{
\includegraphics[scale=0.5]{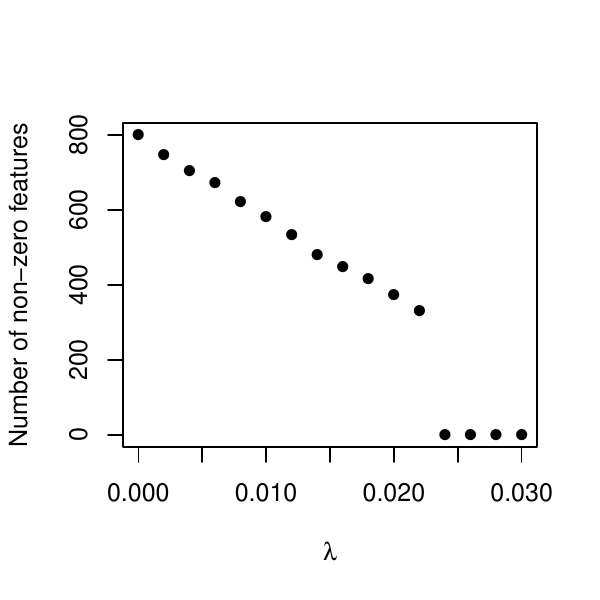}
\includegraphics[scale=0.5]{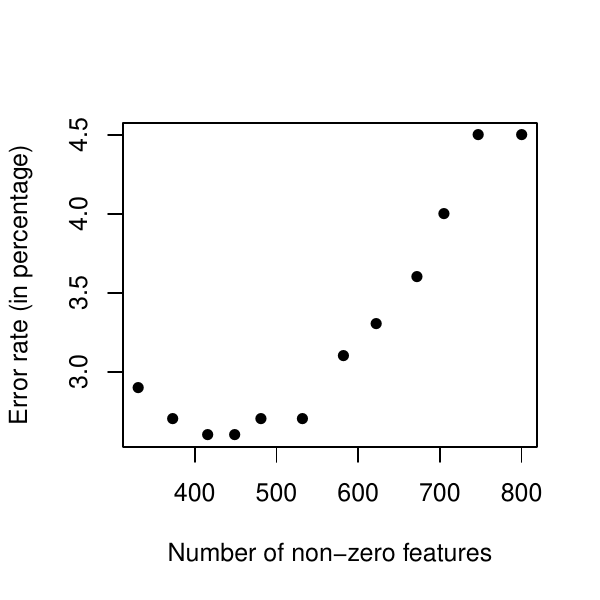}
}
\caption{\label{fig:drop}Relationship between the tuning parameter $\lambda$, sparsity level of the solution $v_{\lambda}$ and the corresponding error rate of penalized FLDA for the Data Based covariance structure.}
\end{figure}

Though FLDA results in good prediction performance, the number of selected features is high in comparison to the number of truly different features. This is not a drawback of cross-validation, but is an intrinsic property of the $\ell_1$ penalized problem. As we discuss in Section~\ref{sec:nonconvexity}, very sparse solutions are not always obtainable. Indeed, we performed a separate study using one of the simulated covariance matrices to evaluate the number of non-zero components selected by the algorithm versus the value of the tuning parameter. The results are shown in Figure \ref{fig:drop}. The drop is observed at around 350 features which is only somewhat less that the mean value for selected features in Table~\ref{tab:simul}. The restriction on the sparsity level, however, does not seem to have a negative effect on the classification performance of the method. Figure~\ref{fig:drop} illustrates how the misclassification rate of FLDA varies with the sparsity level of  $v_{\lambda}$.

\section{Application to DNA Methylation Data}\label{sec:data}

ERASMUS and ECOG datasets described in Section~\ref{sec:intro} have been previously analyzed by \citet{Figueroa:2010ih} and \citet{Kormaksson:V8tzurfm}. These studies revealed that the cluster analysis applied to ERASMUS dataset corresponded well to the available biomarker information, however only limited biomarker information is available for the ECOG dataset. It is of great interest to investigate whether the ERASMUS dataset can be used to identify the leukemia subtypes in the ECOG dataset. For this purpose we apply penalized FLDA with $\tilde W$ estimated according to Section~\ref{sec:penFLDA} to the ERASMUS methylation dataset and use ECOG dataset as a test set.

\begin{figure}
\centering
\includegraphics[scale=0.5]{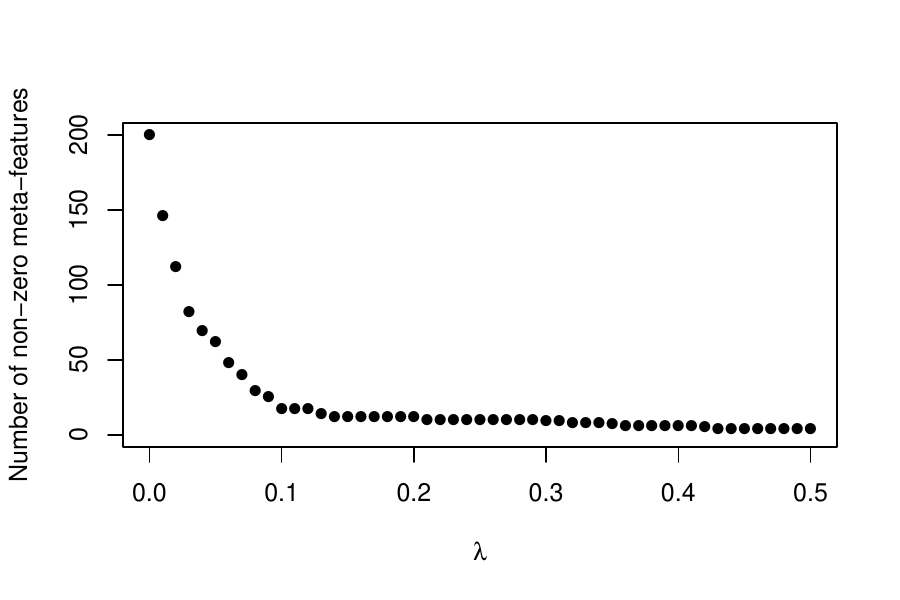}
\caption{\label{fig:clust}Number of non-zero meta-features versus tuning parameter, after clustering the features of ERASMUS data set.}
\end{figure}

Given the large size of the ERASMUS dataset, $p=18954$, it is important to consider sparse solutions of penalized FLDA. While a large number of selected features may not affect the classification performance, it makes the interpretation of the results very challenging. In simulations with Data Based covariance structure in Section~\ref{sec:Simulations} it was not possible to select fewer than 300 features out of 800. This number will be even larger for $p=18954$. Therefore, at first we investigate whether the features clustering idea described in Section~\ref{sec:clustering} improves the sparsity level of solutions obtained by penalized FLDA.  For this purpose we use two biomarkers, t.8.21 (AML1/ETO) and inv.16 (CBFb/MYH11), that are available for both datasets.  Clustering of features is performed using the k-means algorithm based on the difference in sample means between these two biomarkers \citep{Hartigan:1979wo}. The number of clusters is chosen to be $k=200$ (approximately $1\%$ of the number of original features $p=18954$) and the number of random starts is set to 1000. Note that even with such strong dimension reduction, $k$ is still much bigger than $n$ (there are 52 patients with t.8.21 and inv.16 in the ERASMUS dataset). We use these 200 clusters to generate meta-features as discussed in Section~\ref{sec:clustering} and solve optimization problem~\eqref{eq:LDAwithinTilde} for a range of tuning parameters $\lambda$. Figure \ref{fig:clust} shows the relationship between the number of selected meta-features and the value of the tuning parameter. The solution path appears to be smooth in $\lambda$, with larger values of $\lambda$ resulting in very sparse solutions ($<10$ selected meta-features). 

We proceed by analyzing the effect of clustering on the prediction performance. Performing cross-validation on ERASMUS dataset results in selecting the tuning parameter $\lambda=0.24$, which corresponds to five clusters with centers at 1.8, -2.2, -1.7, 2.4 and 1.5. These five selected clusters contain only 79 out of 18954 original features. In contrast, FLDAdiag applied to the same meta-features chooses the tuning parameter that selects 32 clusters (1663 original features). The classification error on the ECOG dataset is zero for both methods. 

To investigate the effect of the number of clusters, $k$, we repeat the analysis with $k=500$. In this case the algorithm selects seven clusters which contain 73 original features. Out of those 73, 51 features are the same as the ones selected with $k=200$. The prediction error on ECOG dataset is again zero. Thus, it appears that the algorithm is robust with respect to the number of clusters $k$.

Given the positive effect of clustering, we repeat the procedure using four biomarkers (inv.16, t.8.21, t.15.17 and double mutants). Since there are 4 subtypes, 3 discriminant vectors can be considered, obtained sequentially as described in Appendix~A. To summarize, the following steps are performed:

\begin{enumerate}
\item Choose the first two subtypes.
\item Cluster the features based on the differences between the two subtypes.
\item Perform discriminant analysis on clusters using only two selected subtypes.
\item Determine which original features correspond to selected clusters and eliminate them from the feature set.
\item Merge two subtypes into one and add one subtype from the remaining ones. 
\item Repeat from Step 2 until the final subtype is added.
\end{enumerate}

\begin{figure}[!h]
\centering
\includegraphics[scale=0.65]{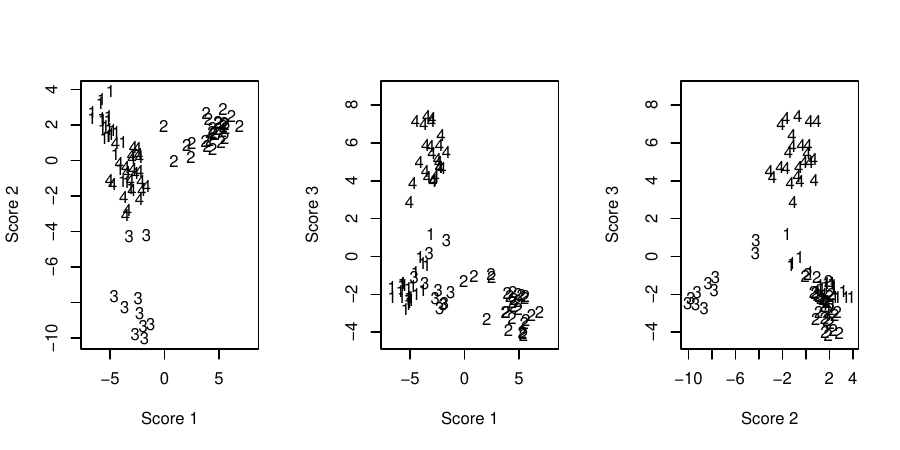}

\caption{Discriminant scores for 4 subtypes projected onto 2 dimensions, ERASMUS.}
\label{fig:ERASMUS}
\includegraphics[scale=0.65]{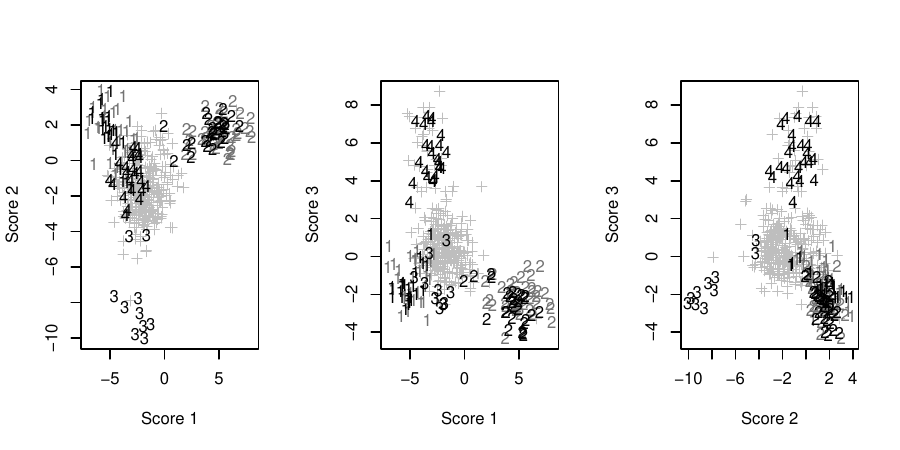}
\caption{Discriminant scores for 4 subtypes projected onto 2 dimensions, ERASMUS+ECOG.}
\label{fig:ECOG}
\end{figure}

The resulting scores for the ERASMUS data are displayed in Figure \ref{fig:ERASMUS}. The resulting scores for the ECOG data plotted on the same coordinates are in Figure \ref{fig:ECOG}. The scores are based on 5, 5 and 7 clusters correspondingly (72, 61 and 104 original features). Symbols ``1'' and ``2'' correspond to subtypes known in both datasets (t.8.21 and inv.16). The figures indicate strong agreement between the ERASMUS and the ECOG datasets in terms  of subtype classification.

\section{Discussion}\label{sec:discus}

In this article we consider an extension of Fisher's discriminant analysis to the case where $p \gg n$. The core of the method is optimization problem \eqref{eq:LDAwithinTilde} which poses several challenges. First, the problem is nonconvex meaning that convergence to the global maximum is not guaranteed. Secondly, the matrices $\tilde W$ and $B$ have to be reliable estimates of their population counterparts $\varSigma_w$ and $\varSigma_b$. Our use of a non-diagonal shrinkage estimator for $\varSigma_w$ not only leads to improved misclassification rates over the method proposed by \citet{Witten:2011kc}, but also makes penalized FLDA competitive with other LDA proposals from the literature as demonstrated in Section~\ref{sec:Simulations}. We also identify problems with selecting a sparse subset of features and propose a solution involving pre-clustering of features.

While in this paper we have restricted our attention to the penalized FLDA problem, we conjecture that the restriction on the solution sparsity is an intrinsic property of an $\ell_{1}$-penalized criterion with nonconvex objective function due to the likely non-zero duality gap. Other examples of such criterions include \citet{Zou:2006ia} and \citet{BIEN:vs}. Such a restriction has direct implications on the variable selection properties of corresponding estimators and in future research we are planning to generalize our results to other non-convex criterions that are used in practice.

\section*{Acknowledgement}

We thank Michael Todd for a useful discussion of duality theory. This research was partially supported by NSF grants DMS-1208488 and DMS-0808864.

\appendix

\section*{Appendix A: Optimization Details}

\subsection*{A.1. Derivation of Algorithm~\ref{a:nondiag}}
Following \citet{Witten:2011kc}, \eqref{eq:LDAwithinTilde} can be recast as a biconvex optimization problem
\begin{equation}\label{eq:biconvex}
 \maximize_{u,v}\left \{2u^{T}B^{1/2}v-\lambda\sum_{j=1}^{p}|s_{j}v_{j}|-u^{T}u\right \} \mbox {subject to } v^{T}\tilde Wv\le1,
\end{equation}
since maximizing with respect to $u$ gives $u=B^{1/2}v$. The problem \eqref{eq:biconvex} is convex with respect to $u$ when $v$ is fixed and is convex with respect to $v$ when $u$ is fixed. This property allows the use of Alternate Convex Search (ACS) to find the solution \citep[Section~4.2.1]{Gorski:2007uv}. ACS ensures that all accumulation points are partial optima and have the same function value \citep[Theorem~4.9]{Gorski:2007uv}.

Starting with an initial value $v^{(0)}$ the algorithm proceeds by iterating the following two steps:
\begin{description}
  \item[Step 1] $u^{(k)}=\arg\max_u\left\{2u^{T}B^{1/2}v^{(k)}-u^{T}u\right\}=B^{1/2}v^{(k)}$
 \item[Step 2] 
 $v^{(k+1)}=\arg\max_v\left\{2(u^{(k)})^{T}B^{1/2}v-\lambda\sum_{j=1}^{p}|s_{j}v_{j}|\right\} \mbox {subject to } v^{T}\tilde Wv\le1.$
\end{description}

The main challenge is to solve Step 2. Following \citet[Proposition~2]{Witten:2011kc}, it is useful to reformulate Step 2 as
\begin{equation}\label{eq:LDAk}
q^{(k+1)}=\arg \max_q\left \{2(u^{(k)})^{T}B^{1/2}q-\lambda\sum_{j=1}^{p}|s_{j}q_{j}|-q^{T}\tilde Wq\right \}
\end{equation}
where, if $q^{(k+1)}=0$, then $v^{(k+1)}=0$, else $v^{(k+1)}=q^{(k+1)}/\sqrt{(q^{(k+1)})^{T}\tilde Wq^{(k+1)}}$. Since problem \eqref{eq:LDAk} is convex with respect to $q$, the solution $q^{(k+1)}$ satisfies KKT conditions \citep{Boyd:2004uz}
\begin{equation}\label{eq:KKT}
 2B^{1/2}u^{(k)}-2\tilde Wq^{(k+1)}-\lambda \varGamma=0,
\end{equation}
where $\varGamma$ is a $p$-vector and each $\varGamma_j$ is a subgradient of $|s_{j}q_j|$, i.e. $\varGamma_{j}=s_j$ if $
q^{(k+1)}_{j}>0$, $\varGamma_{j}=-s_j$ if $q^{(k+1)}_{j}<0$ and $\varGamma_{j}$ is between $-s_j$ and $s_j$ if $q^{(k+1)}_{j}=0$.
Solving \eqref{eq:KKT} with respect to individual components of the vector $q^{(k+1)}$ gives
\begin{equation}\label{eq:qupdate}
q^{(k+1)}_{j}=\frac{S\left((B^{1/2}u^{(k)})_{j}-\sum_{i\ne j}w_{ji}q^{(k+1)}_{i},\frac{\lambda s_j}2\right)}{w_{jj}}.
\end{equation}
Because of the additional term, $\sum_{i\ne j}w_{ji}q^{(k+1)}_{i}$, each component of vector $q^{(k+1)}$ depends on the other components and therefore the solution is not available in closed form. We propose to use coordinate update to overcome this challenge.
Let $$f(q)=2u^{T}B^{1/2}q-\lambda\sum_{j=1}^{p}|s_{j}q_{j}|-q^{T}\tilde Wq$$ be the objective function and note that $f$ can be written as $f(q_1,...,q_p)=f_0(q_1,...q_p)+\sum_{j=1}^p f_j(q_j),$ where $f_0(q_1,..,q_p)=2u^{T}B^{1/2}q-q^{T}\tilde Wq$ is concave and differentiable in $q$ and $f_j(q_j)=-\lambda|s_j q_j|$ is concave in $q$ for each $j$. It was established by \citet{Tseng:1988ut} that coordinate ascent methods converge for such functions. Algorithm~\ref{a:nondiag} results from combining Steps 1 and 2 with sequential update~\eqref{eq:qupdate}.

\subsection*{A.2. Selection of the tuning parameter $\lambda$}\label{lambda}
It is traditional to choose the tuning parameter by cross-validation: the final $\lambda$ is chosen from the  respective grid $\lambda_{1}\le...\le\lambda_{\max}$ to minimize the cross-validation error rate. In LASSO \citep{Tibshirani:1996wb}, $\lambda_{\max}$ results in a solution vector that is exactly zero. However it is not clear what is the appropriate value for $\lambda_{\max}$ in the FLDA context. Here we provide some intuition on how this value is chosen.
  
From the form of the update on $q$ given in Algorithm~\ref{a:nondiag}, for a given iteration $k$ all components of $v^{(k)}$ are set to zero if and only if 
$ \max_j \left|(2Bv^{(k-1)})_j/s_j\right|\le \lambda.$
It follows that if $v^{(k-1)}=0$, then all subsequent $v^{(k)}=0$ and the algorithm terminates. As a special case, $v^{(1)}=0$ if and only if
$ \max_j \left|(2Bv^{(0)})_j/s_j\right|\le \lambda,$
where $v^{(0)}$ is the initial value of $v$. This leads to an upper bound on $\lambda$:
\begin{equation*}
\lambda_{max}=2\max_j \left|(Bv^{(0)})_j/s_j\right|.
\end{equation*}
This bound depends on the $v^{(0)}$, which is natural since $v^{(0)}$ corresponds to the solution of Fisher's discriminant problem without the $\ell_1$ penalty. 

\subsection*{A.3. Producing several discriminant vectors}\label{s:SevVect} 

Previously we have only considered the problem of finding the first discriminant vector $v=v_{1}$, however when $g>2$ additional discriminant vectors are often of interest. \citet{Witten:2011kc} address this problem by finding discriminant vectors sequentially. The $l$th discriminant vector $v_{l}$ is found as the solution to optimization problem \eqref{eq:LDAwithinTilde} with the matrix $B$ replaced by matrix $B^{l}$. Let $X$ be an $n \times p$ standardized data matrix and let $Y$ be an $n \times g$ group indicator matrix where $Y_{ig}=1$ if observation $i$ is in group $g$ and 0 otherwise. Then $B^{l}$ is defined as
$$B^{l}=X^{T}Y(Y^{T}Y)^{-1/2}P_l^{\perp}(Y^{T}Y)^{-1/2}Y^{T}X,$$
where $P_1^{\perp}=I$ and $P_l^{\perp}$ is an orthogonal projection matrix into the space orthogonal to $(Y^{T}Y)^{-1/2}Y^{T}Xv_{i}$ for all $i<l$. Note that $Y^TY$ is a $g \times g$ diagonal matrix where each diagonal element is the number of observations in the corresponding group and $Y^{T}X$ sums the features of $X$ by group. 

Though this approach guarantees orthogonality of discriminant vectors, it also provides additional challenge in interpretation since the same features may be involved in more than one vector. A simpler approach is to find discriminant vectors by reducing the feature space sequentially. After the first discriminant vector is produced, the features corresponding to its non-zero components are eliminated and then the algorithm is repeated on the remaining features.

\section*{Appendix B: Proofs of Propositions~\ref{cl:dualone}-\ref{cl:main}}
\label{sec:Appendix}

\begin{proof}[Proof of Proposition~\ref{cl:dualone}]
Fix any $\lambda \ge 0$ and let $v_{\lambda}$ be the corresponding solution of \eqref{eq:LDAl}. Then it follows that for any $v$ such that $v^T\tilde Wv\le 1$
\begin{equation}\label{eq:vol}
v_{\lambda}^TBv_{\lambda}-\lambda\|v_{\lambda}\|_1 \ge v^TBv-\lambda \|v\|_1.
\end{equation}
Consider \eqref{eq:LDAt} with $t=\|v_{\lambda}\|_1$.  From \eqref{eq:vol} for each $v$ such that $v^T\tilde Wv\le 1$ and $\|v\|_1\le t$ 
\begin{equation*}
v_{\lambda}^TBv_{\lambda}\ge v^TBv+\lambda(\|v_{\lambda}\|_1-\|v\|_1)=v^TBv+\lambda(t-\|v\|_1)\ge v^TBv.
\end{equation*}
This means $v_{\lambda}$ is the solution to \eqref{eq:LDAt}, hence $v_{t}=v_{\lambda}$.
\end{proof}

\begin{proof}[Proof of Proposition~\ref{cl:ttest}]
For any vector $z\in \mathbb{R}^p$, $B'z=\gamma ll^{T}z$, so $(B'z)_i=\gamma(l^Tz)l_i$ and $\sign((B'z)_i)=\sign((l^Tz)l_i)$. Therefore $|(B'z)_i|\ge|(B'z)_j|$ iff $|l_i|\ge |l_j|$. Since the algorithm update for solving \eqref{eq:zLDAs1} has the form
$$
z^{(k+1)}_j=\frac{\sign((Bz^{(k)})_j)\max(0,|(Bz^{(k)})_j|-\lambda/2)}{\delta},
$$
it follows that if $|z^{(k+1)}_k|>0$ then $|z^{(k+1)}_i|>0$ for all $i<k$ since $|l_1|\ge|l_2|\ge...\ge|l_p|>0$. Hence, non-zero elements of $z_{\lambda}$ correspond to the largest values of $|l_{i}|$. Since $l\propto t$ with $t_{i}=(\bar X_{1i}-\bar X_{2i})/s_{i}$, the result of the proposition follows.
\end{proof}

 \begin{lemma}\label{cl:bounds} Let $f(z)=z^{T}B'z-\lambda\|z\|_1$. Define $F_0=f(0)=0$; $F=\max_{v^Tv\le1} f(v)=f(v_{\lambda})$ and
$$F_j=\max_{v \in A_j\cup A_0} f(v) \mbox{ subject to } v^{T}v\le 1.$$
Then for each $j=1,...,p$: $\underline{F_j}\le F_j\le \overline{F_j}$, where
$$\underline{F_j}=\gamma \|l^j\|^2_2-\lambda\frac{\|l^{j}\|_1}{\|l^{j}\|_2},\quad \overline{F_j}=\max\left(0, \gamma \|l^j\|^2_2-\lambda\frac{\|l^{j}\|_2}{|l_1|}\right).$$
\end{lemma}
\begin{proof}[Proof of Lemma~\ref{cl:bounds}]
1. Show $\underline{F_j}\le F_j$. By definition $F_j\ge f(v')$ for all $v'\in A_k \cup A_0$ with $v'^Tv'\le 1$. Take $v'=l^j/\|l^j\|_2$. Then $F_{j}\ge f(v')$ and
$$
f(v')=v'^{T}\gamma ll^{T}v'-\lambda\sum_{i=1}^{p}|v'_i|=\gamma (l^{T}v')^2-\lambda\sum_{i=1}^{j}|v'_i|=\underline{F_j}.
$$
2. Show $F_j\le \overline{F_j}$. For all $v \in A_j\cup A_0$ with $v^Tv\le 1$ 
\begin{equation*}
\begin{split}
f(v)&=\gamma (l^{T}v)^2-\lambda\sum_{i=1}^{j}|v_i|=\gamma\sum_{i=1}^j|l_i| |v_i|\left(\sum_{i=1}^j|l_i| |v_i|-\frac{\lambda}{\gamma|l_i|}\right)\\ 
&\le \gamma\sum_{i=1}^j|l_i| |v_i|\left(\|l^j\|_2-\frac{\lambda}{\gamma|l_1|}\right) \le \max\left(0,\gamma \|l^j\|_2\left(\|l^j\|_2-\frac{\lambda}{\gamma|l_1|}\right)\right)=\overline{F_j}.
\end{split}
\end{equation*}
\end{proof}

\begin{proof}[Proof of Proposition~\ref{cl:additional}]
If $F_j\le0$ then $v_{\lambda}\notin A_j$ since $F_0=0\ge F_j$ and $F=\max(F_0,F_1,...,F_p)$. From here and Lemma~\ref{cl:bounds} it follows that a necessary condition for $v_{\lambda} \in A_j$ is that $\overline{F_j}>0$. Indeed, if $\overline{F_j}=0$ then $F_j\le \overline{F_j}= 0$. By definition $\overline{F_j}>0$ is equivalent to $\gamma |l_{1}| \|l^j\|^2_2-\lambda\|l^{j}\|_2>0.$
This can be rewritten as $\|l^j\|_2(\gamma|l_{1}|\|l^j\|_2-\lambda)>0$, which is equivalent to $\|l^j\|_2>\lambda/(\gamma |l_1|)$. Since $\|l^j\|_2\le\|l^m\|_2$ for all $m\ge j$, this means that if $\overline{F_j}>0$ then $\overline{F_m}>0$ for all $m\ge j$. On the other hand, if $\overline{F_m}=0$ then $\|l^m\|_2\le \lambda/(\gamma|l_1|)$, which means that $\overline{F_j}=0$ for all $j\le m$. It follows that if $\overline{F_m}=0$ then $v_{\lambda}\notin \bigcup_{i=1}^{m}A_i$ which is equivalent to $v_{\lambda} \in A_0\cup \bigcup_{i=m+1}^pA_i$. 
\end{proof}

\begin{proof}[Proof of Proposition~\ref{cl:main}]
A sufficient condition for $F_r>F_k$ when $r>k$ is $\underline{F_r}>\overline{F_k}$. If $\overline{F_k}=0$ then $v_{\lambda}\notin \bigcup_{i=1}^{k}A_i$, therefore it is enough to consider the case $\overline{F_k}>0$. From Proposition \ref{cl:bounds} this is equivalent to
$$
\gamma\|l^r\|^2_2-\lambda\frac{\|l^r\|_2}{\|l^r\|_1}>\gamma\|l^k\|^2_2-\lambda\frac{\|l^k\|_2}{|l_1|},
$$
which can be rewritten as
$$
\frac{\lambda}{\gamma}<\left(\|l^r\|^2_2-\|l^k\|^2_2\right)/\left(\frac{\|l^r\|_2}{\|l^r\|_1}-\frac{\|l^k\|_2}{|l_1|}\right).
$$
We know that $\overline{F_k}>0$ iff $\lambda<\gamma|l_1|\|l^k\|_2$. It follows that $\underline{F_r}> \overline{F_k}>0$ is equivalent to
$$
|l_1|\|l^k\|_2\le \left(\|l^r\|^2_2-\|l^k\|^2_2\right)/\left(\frac{\|l^r\|_2}{\|l^r\|_1}-\frac{\|l^k\|_2}{|l_1|}\right).
$$
This leads to $ \|l^k\|_2\le \|l^r\|^{3/2}/(|l_1|\|l^r\|_1).$
Since $\|l^m\|_2\ge \|l^k\|_2$ for all $m\ge k$, the result follows.
\end{proof}

\bibliographystyle{biometrika}
\bibliography{Rpackage,PaperDraft}
\end{document}